\documentclass{article} 
\PassOptionsToPackage{authoryear}{natbib}
\usepackage[final]{nips_2017}
\usepackage{times,amsmath}
\usepackage{url}
\usepackage{color}
\usepackage{graphicx}
\usepackage{subcaption}

\title{Safer Classification by Synthesis}

\author{William Wang, Angelina Wang, Aviv Tamar, Xi Chen \& Pieter Abbeel\\ 
Dept. of Electrical Engineering and Computer Sciences\\
UC Berkeley\\
Berkeley, CA 94709
}

\DeclareMathOperator*{\argmax}{arg\,max}

\begin{document}
\maketitle
\begin{abstract}
The discriminative approach to classification using deep neural networks has become the de-facto standard in various fields. Complementing recent reservations about safety against adversarial examples, we show that conventional discriminative methods can easily be fooled to provide incorrect labels with very high confidence to out of distribution examples. We posit that a generative approach is the natural remedy for this problem, and propose a method for classification using generative models.
At training time, we learn a generative model for each class, while at test time, given an example to classify, we query each generator for its most similar generation, and select the class corresponding to the most similar one. Our approach is general and can be used with expressive models such as GANs and VAEs. At test time, our method accurately ``knows when it does not know,'' and provides resilience to out of distribution examples while maintaining competitive performance for standard examples. 
\end{abstract}
\section{Introduction}

\begin{quote}
    ``What I cannot create, I do not understand."
\end{quote}
This famous quote by Richard Feynman stands in stark contrast to the majority of image classification breakthroughs of the last decade. The prevalent deep learning approach is \emph{discriminative}, where a deep network maps an observation into a probability over decisions, providing little to no understanding of why a particular decision is chosen (e.g.,~\citealt{krizhevsky2012imagenet,szegedy2015going,he2016deep} among others). While this approach has demonstrated remarkable empirical results, it’s opaque nature raises questions of safety and trust~\citep{amodei2016concrete}.

For example, much attention has recently been focused on dealing with adversarial perturbations to discriminative image classifiers, which make small image modifications that result in misclassification~\citep{goodfellow2014explaining}. In this work we complement this view, by showing that discriminative models can also be easily fooled to give erroneous predictions \emph{with a high confidence} for out-of-distribution examples, which are significantly different from examples in the data. 

Motivated by the safety issues of discriminative classifiers, in this work we propose a safer \emph{generative} image classification paradigm. We build on recent breakthroughs in deep generative modelling such as variational autoencoders (VAEs; \citealt{kingma2013auto}) and generative adversarial nets (GANs; \citealt{goodfellow2014generative}), which have shown convincing results for generating complex observations such as images directly from data. 
Our idea is to use labeled training data for building generative models for images from each class. In these models (either GANs or VAEs), a random input vector is transformed by a deep neural network into an image. At test-time, given an image to classify, we \emph{search} across the input vectors for an image that is sufficiently similar to the test image, across all the generators. The corresponding class of the best image is taken as the classification result. 

An immediate benefit of this approach is interpretability -- once a class decision is made, we know exactly why it was chosen, since \emph{we readily know the image which our model imagined as most representative} of it. 
Another benefit, which we show here to be significant, is safety. When a traditional discriminative model is trained to classify, say, road signs, we have no idea what it would do when faced with an out-of-distribution example, say, of an elephant. This poses a severe problem for safety critical systems such as self driving cars. Our model, by definition, would never generate an image of an elephant if elephant images are not in the training data. Thus, \emph{distinguishing when the model does not know becomes straightforward}.

This brings us back to the premise in the Feynman quote. For objects that we understand, and can therefore reliably generate, our model provides reliable classification. 










\subsection{Related Work}
Generative classification is an old idea. Using shallow architectures, \cite{ng2002discriminative} compared discriminative and generative learning by investigating logistic regression and naive Bayes, and observed improved performance of generative models in low-data regimes. \cite{jaakkola1999exploiting} used a generative model to extract kernels for a discriminative logistic regression classifier.

In the seminal work of \citet{hinton2006fast} on deep belief networks (DBNs), deep generative models for images and matching class labels were learned. For classification using DBNs, the image is used to calculate activations of a restricted Boltzman machine (RBM) for the image and label, which are clamped, and MCMC sampling is then used to generate the corresponding label. Over the last decade, DBNs have been outperformed by discriminative models trained using backpropagation~\citep{vincent2010stacked,krizhevsky2012imagenet}. 





More recently, generative models that can be trained using backpropagation have become popular. VAEs~\citep{kingma2013auto} and GANs~\citep{goodfellow2014generative} can be seen as extensions of the Helmholtz machine model~\citep{dayan1995helmholtz}, where a random vector with a known distribution is mapped through a neural network to generate the data distribution. VAEs can be trained using a variational lower bound, while GANs are trained using an adversarial method. While such generative models have been used with success in semi-supervised learning \citep{kingma2014semisupervised}, they have not yet been explored in the context of supervised learning.
Given a test image, the VAE recognition network can be used to sample from the distribution over latent variables, and generate a similar image. 
For GANs, models that learn an inference network have been proposed~\citep{donahue2016adversarial,dumoulin2016adversarially}.
In this work we describe an alternative inference approach for GANs that does not require a change to the training objective.

Selective classification~\citep{chow1957optimum,el2010foundations} is an established approach for improving classification accuracy by rejecting examples that fall below a confidence threshold. In this work we use this method for rejecting out-of-distribution examples based on the confidence score. We build on the recent work of \citet{geifman2017selective}, who investigated suitable confidence scores for deep neural networks. Very recently, \citet{mandelbaum2017distance} studied different distance metrics for use as a confidence score.

In concurrent work by other authors, generative models have been used to detect examples that are outliers or outside the training distribution.~\cite{anonymous2018anomaly} uses a similar technique of searching over the latent space of a GAN to discover outliers, but does not use such a technique for classification.~\cite{anonymous2018novelty} trains GANs with feature matching loss to develop a model that is capable of simultaneous classification and novelty detection. While both of these use ROC curves to benchmark their model, in our context of classification, we believe that the risk-coverage analysis is more appropriate, as it evaluate the confidence of both classification performance and novelty detection. An ROC analysis of novelty detection, on the other hand, ignores information about the classification accuracy.




\subsection{Our Contribution}
In this work we make the following contributions.
\begin{enumerate}
    \item Using the selective classification paradigm, we show in a principled way that discriminatively trained deep neural networks can easily be fooled by out-of-distribution examples.
    \item We propose a general method for generative classification, that can be used with either GANs, VAEs, or even K-nearest neighbor methods as the generative model.
    \item We show that our method can provide significantly better resilience to out-of-distribution examples, while maintaining competitive accuracy on within-class examples. 
\end{enumerate}

\section{Preliminaries}
Consider a classification problem with $K$ classes. The inputs (e.g., images) are denoted by $x$ and the outputs (classes) are denoted by $y$. Our data consists of pairs $\left\{ x_i, y_i\right\}$, where $x_i\sim P(x)$ and $y_i \sim P(y|x_i)$. 

Discriminative classification learns a model $P_\theta(y|x)$ with parameters $\theta$, typically by maximizing the data log-likelihood $\max_\theta \sum_i \log P_\theta(y_i|x_i)$. A popular model for classification is the softmax $P_\theta(y=k|x) = \frac{\exp{f_{\theta,k}(x)}}{\sum_j\exp{f_{\theta,j}(x)}}$, where $f_{\theta,k}(x)$ for $k=1,...,K$ are a deterministic functions, e.g., the outputs of a neural network with weights $\theta$.

\subsection{Generative Models}
The central component of our method is a \textit{generative mo
del} $G$, which takes as input a random $m$-dimensional latent vector $z$ from a distribution $P(z)$ and learns to transform $z$ into a sample from $P(x)$. Learning $G$ is an unsupervised learning task, and recently, several  efficient training methods that use backpropagation were proposed, such as GANs~\citep{goodfellow2014generative} and VAEs~\citep{kingma2013auto}.

GANs are adversarially trained networks consisting of a generator and a discriminator. The generator network takes a latent vector and produces an image, while the discriminator network takes an image and outputs a predicted probability that this image came from the distribution of the training set. Training consists of a two-player game where the generator tries to produce images that the discriminator is unable to distinguish from the training distribution, while the discriminator improves its ability to discern between real data and generated images. 

VAEs are trained using a variational lower bound, by learning a encoder network that maps a training image into a corresponding distribution over the latent vector, and a decoder network that maps the latent vector back into an image. The training balances the reconstruction loss of the decoder with the Kullback-Leibler distance of the encoded latent vector distribution from $P(z)$.

\subsection{Selective Classification}
A central motivation for our approach is accurately identifying when the classifier `does not know' the correct class. This problem, typically explored in the context of discriminative classification, is known as selective classification~\citep{geifman2017selective}.

Suppose we have a classifier $f$ which takes as input an image and outputs a predicted class along with a measure of confidence in its prediction. For example, $f$ could be a CNN with the maximum softmax output as its confidence. \textit{Selective classifiers} abstain from prediction if the confidence score is below a certain threshold $\theta$. 

\begin{figure}
    \centering
    \includegraphics[scale=.6]{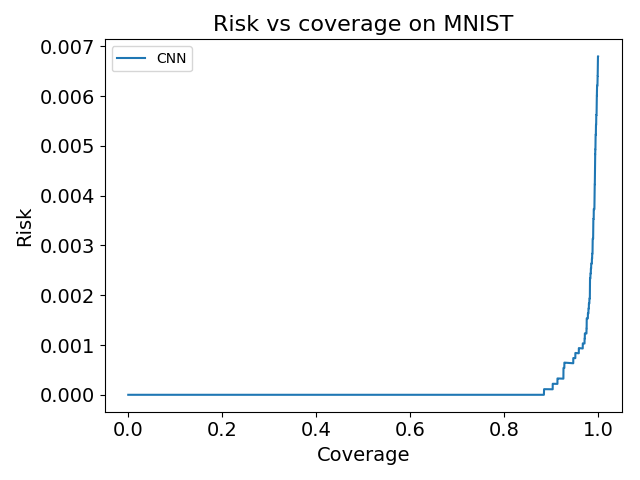}
    \caption{Risk-coverage plot for a CNN trained on MNIST. Observe that we can choose a confidence threshold such that around 90\% of the data is covered, and for this data portion the classification is perfect. As we increase the desired coverage, the accuracy decreases (risk increases). Choosing a threshold that covers more than 99 percent of the data will result in a sharp decrease in accuracy, which indicates that the low-confidence predictions generally correspond to misclassified data, as we would expect.}
    \label{cnn_test}
\end{figure}

The threshold parameter thus offers a balance between the proportion of the data classified and the accuracy on this portion of the dataset.
The coverage of a selective classifier with threshold $\theta$ is defined as the proportion of test observations that are classified with confidence greater than $\theta$ \citep{el2010foundations}. The empirical risk given $\theta$ is then defined as the error rate on the subset of the test set that was classified with confidence greater than $\theta$.
A principled method for comparing selective classifiers is to examine their \emph{risk-coverage plots}, as exemplified in Figure~\ref{cnn_test}.
Classifiers with meaningful measures of confidence should predict difficult or out of distribution images with lower confidence, and thus, \emph{as the coverage is decreased, the risk should shrink to zero}.

Recently, \cite{geifman2017selective} showed that for discriminatively-trained CNNs, thresholding the softmax output provides state-of-the-art selective classification, surpassing alternative confidence measures such as MC-Dropout~\citep{gal2016dropout}. 

\section{Discriminative Models are Easily Fooled by Out-of-Distribution Examples}

We begin our discussion by showing that a discriminatively trained CNN can easily be fooled to give high-confidence predictions to out-of-distribution examples -- examples that are significantly different from any example in its training data, and that do not correspond to any particular class it was trained to predict.

We present our results under the selective classification paradigm, for a principled evaluation of confidence and accuracy. Consider the classification task where a selective classifier is trained on data containing a certain set of classes, but is tested on data $X_1\cup X_2$ where $X_1$ contains data from classes seen during training and $X_2$ contains data that do not match any of the classes seen in the training set. Obviously, the classifier will be unable to correctly classify any points in $X_2$, so we expect it to abstain from prediction on these points. Concretely, we would like to pick a threshold $\theta$ such that most points in $X_2$ have confidence less than $\theta$ and so are left unclassified, while most points in $X_1$ have confidence greater than $\theta$ and are classified correctly. 
If the data in $X_1$ and $X_2$ is significantly different and our confidence measure is reliable, we should be able to determine such a threshold.

Unfortunately, as we show next, discriminative CNN classifiers are easily fooled to give high-confidence predictions for out-of-distribution examples that are \emph{wildly different} from their training data.

\subsection{MNIST Augmented with Omniglot}

\begin{figure}[!h]
    \centering
    \begin{subfigure}{0.23\textwidth}
        \centering
        \includegraphics[scale=.2]{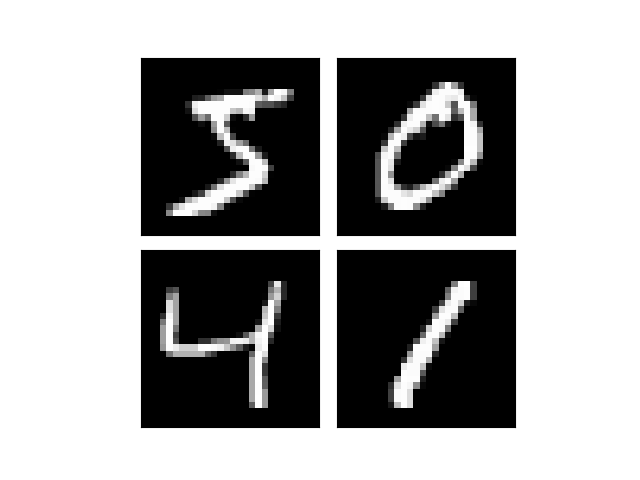}
        \caption{MNIST}
        \label{mnist_examples}
    \end{subfigure}
    \hspace{1em}
    \begin{subfigure}{0.23\textwidth}
        \centering
        \includegraphics[scale=.2]{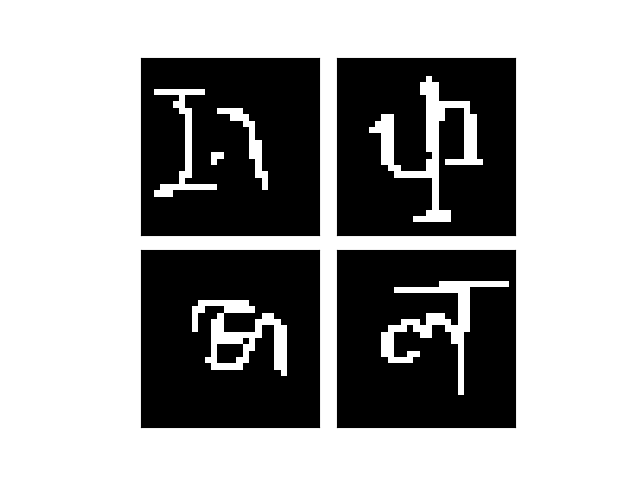}
        \caption{Omniglot}
        \label{omniglot_examples}
    \end{subfigure}
    \begin{subfigure}{0.23\textwidth}
        \centering
        \includegraphics[scale=.2]{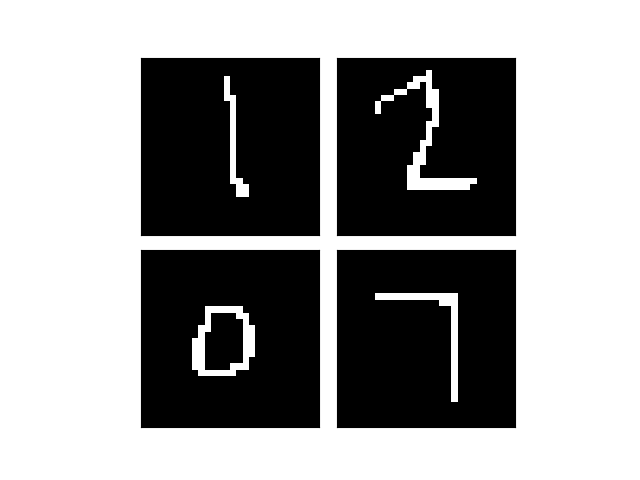}
        \caption{Removed Omniglot}
        \label{removed_omniglot_examples}
    \end{subfigure}
    \caption{MNIST and Omniglot data}
    \label{data_images}
\end{figure}

We train a standard CNN on the well-known MNIST dataset and test its selectivity by running predictions on MNIST augmented with 31460 rescaled images from the Omniglot dataset. The Omniglot dataset, compiled by \cite{lake2015omniglot}, contains handwritten characters from 50 different alphabets, such as in Figure \ref{omniglot_examples}. The images were resized with nearest neighbor interpolation in order to be the same 28 by 28 size as MNIST. We removed some ambiguous images such as those in Figure \ref{removed_omniglot_examples}, where characters from different languages resembled digits to the point that even a human would categorize them as numerical digits. Our augmented dataset is composed of 24 percent MNIST data and 76 percent Omniglot. Because the Omniglot images we chose to include do not resemble the images from the MNIST training set, we should expect the CNN to predict these images with low confidence. 
Thus, we expect the risk-coverage plot\footnote{For calculating the risk in the risk coverage plot with the augmented data set, every prediction of an Omniglot image is considered to be wrong.} to be similar to Figure~\ref{cnn_test} in that it starts off low and increases once the images the CNN is uncertain about are included in classification. Because MNIST composes 24 percent of our data and the CNN performs well on the MNIST dataset, we expect a flat line that begins to rise monotonically after the 24 percent coverage mark. Further, we expect the risk to decrease towards 0 as we decrease the coverage - this ``vanishing risk'' property reflects the idea that increased confidence should be associated with increased classification accuracy, and can be observed in the MNIST experiment in Figure~\ref{cnn_test}.


\begin{figure}
    \centering
    \includegraphics[width=.6\textwidth]{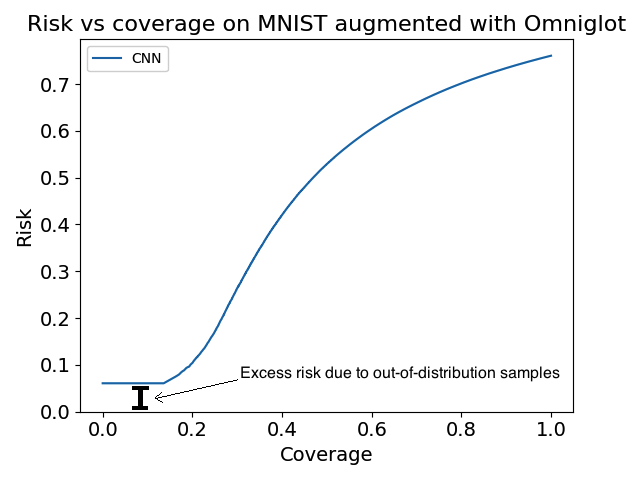}
    \caption{Risk-coverage for CNN on augmented dataset. The lowest attainable risk is .061.}
    \label{cnn_omniglot}
\end{figure}
\begin{figure}
    \centering
    \includegraphics[scale=.3]{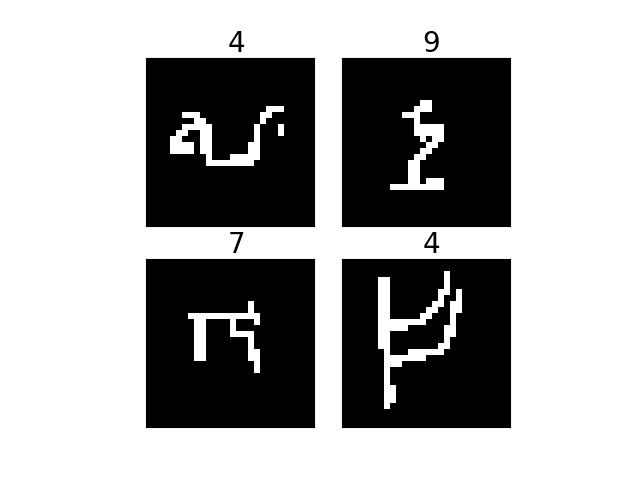}
    \caption{Omniglot Images Classified by CNN with Highest Confidence}
    \label{cnn_omniglot_confident}
\end{figure}

However, the risk-coverage plot we actually obtained for this experiment, shown in Figure~\ref{cnn_omniglot}, does not exhibit the vanishing risk property. 
The lowest risk attainable by the CNN is .061 using the maximum possible confidence threshold of 1 (up to floating point precision) -- in this case, 343 of the 5644 images classified with this threshold of confidence were Omniglot images. These images, displayed in Figure \ref{cnn_omniglot_confident}, attain the highest level of confidence despite not resembling any digit from the MNIST training set, and thus there is no choice of threshold that will allow the CNN to abstain from classifying these images. Combining the risk-coverage results from the MNIST and augmented MNIST datasets, we see that if an image is classified with low confidence, then it is likely an incorrect classification, but the converse is not true: if an image would be incorrectly classified, then there is still a good chance that its prediction confidence was high. Thus the CNN confidence metric does not accurately reflect the ability of the classifier to make an precise prediction, given out-of-distribution examples.

The results on this toy example may seem innocuous at first glance. However, one can easily imagine a scenario where such performance would lead to dire consequences. For example, a realistic scenario for self-driving cars is to defer a decision about road signs to a human based on its confidence in prediction. 
From what we learned in this toy example, a CNN confidence cannot be trusted if, say, a new road sign is introduced, but also if any random object not seen during training is encountered on the road. 

\begin{figure}[h]
    \centering
    \includegraphics[width=.5\textwidth]{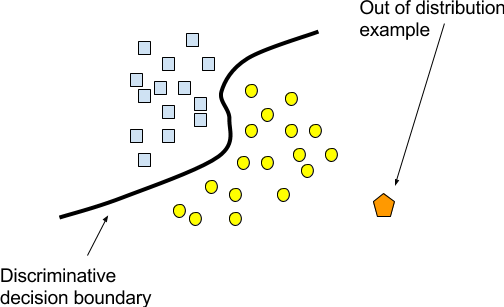}
    \caption{An illustration of an out-of-distribution example with a discriminative classifier. Here, a classifier discriminates between the square and circle examples. We can relate the confidence of the classifier with the distance to the decision boundary. As shown, an out-of-distribution example that is far away from this boundary will be classified with a high confidence to belong, in this case, to the class of circles.}
    \label{intuitive_example}
\end{figure}

In principle, the fact that a discriminative approach can be fooled by out-of-distribution examples should not be very surprising. In Figure \ref{intuitive_example} we provide an explanation for this result in a simple binary classification task. Intuitively, the confidence of a discriminative classifier can be related to the distance from the decision boundary. Therefore, we can imagine that there exists examples that are very different from our data, but still lie far away from the decision boundary, and therefore have a high confidence value. While it is not immediate that the conclusion from this toy example carries over to high-dimensional problems and expressive CNN classifiers, our results above suggest that this is indeed the case. 

From Figure~\ref{intuitive_example} it is also clear that a viable solution for the out-of-distribution detection problem is to identify examples that are too far, in some suitable distance metric, from the training data~\citep{mandelbaum2017distance}. The problem then becomes how to identify a suitable distance metric, and how to compute the distance efficiently, as typically the distance computation scales with the amount of training data.
In the following we propose an alternative approach based on generative models. The idea is that by learning to generate samples from a low-dimensional latent vector, we would effectively learn the manifold for each class. The distance to each manifold is expected to be a reliable measure for classification confidence.

\section{Generative Approach to Classification}
We now propose a different approach to classification, which, by relying on a generative model, provides a better signal for knowing when an example from an unknown class is encountered.

\subsection{Generative Classifier Model}
Our generative classifier consists of class-conditional generative models $G_k(z)$ and a similarity measure $s(x_1,x_2)$. Each generator takes as input a latent variable (say, a random uniform variable over $[-1,1]^m$) and outputs a generated image of its respective class. The similarity measure $s(x_1,x_2)$ could be the negative $L^2$ or $L^1$ distance between $x_1$ and $x_2$\footnote{These similarities would be the negative distance, since according to our convention, larger values of $s(x_1,x_2)$ should mean the images are more similar.}, or it could be a more complex function such as a Siamese network \citep{koch2015siamese} that predicts the probability the two images are of the same class.

In this work we only consider the negative $L^2$ distance, which allows a fair comparison with conventional novelty detection approaches that use similarity metrics. In the future we will investigate using alternative measures of similarity not exclusive to metrics.


To classify a test image $x$ with class $y$, for each generator $G_k$, we solve the following optimization problem:
\begin{equation}\label{optimization_problem}
    z_k^*=\argmax_z s(x,G_k(z))
\end{equation}
That is, for each class $k$, we find the most similar image in the range of $G_k$ to the test image $x$ under the similarity measure $s$, and keep track of the latent vector that produces it, $z_k^*$. Once we have the optimal latent vectors for each class, we classify $x$ as
$$\argmax_k s(x, G_k(z_k^*))$$
In practice, (\ref{optimization_problem}) is a non-convex optimization problem. As an optimization heuristic, we perform Monte Carlo sampling from $P(z)$, evaluate the similarity to the generated image for each latent variable, and use the optimal latent vector as a starting point for a further non-linear optimization method such as L-BFGS~\citep{nocedal1980updating}. If the generative model is a VAE with a Gaussian latent model, then we also have the option of feeding the test image into the encoder network to obtain a parameter estimate for the mean of the Gaussian and use that as a heuristic to solve the optimization. We find that this method works better in practice for the goal of classification accuracy.
\subsection{Comparison with Nearest Neighbors}
Our approach has parallels with the 1-nearest neighbor classifier - in both methods, to classify a test point, there is an optimization performed to find the most similar match over some set of images with known classes. For nearest neighbors, this is the training set, while for our generative classifier, this is the set of all images that are in the range of our generators. Because the generators are presumably capable of reproducing the training set, we would expect our method to outperform nearest neighbors in classification accuracy, as long as the generator spaces for each class do not intersect. Nearest neighbors can be improved on by using different distance metrics - for example, the $L^3$ distance is known to outperform the $L^2$ distance on MNIST. Similarly, it would be possible to use such distance metrics with our generative classifier. While the runtime of nearest neighbors increases with the number of training samples, our method does not, and its runtime is controlled by hyperparameters for the optimization routine. 

\begin{figure}
    \centering
    \includegraphics[width=\textwidth]{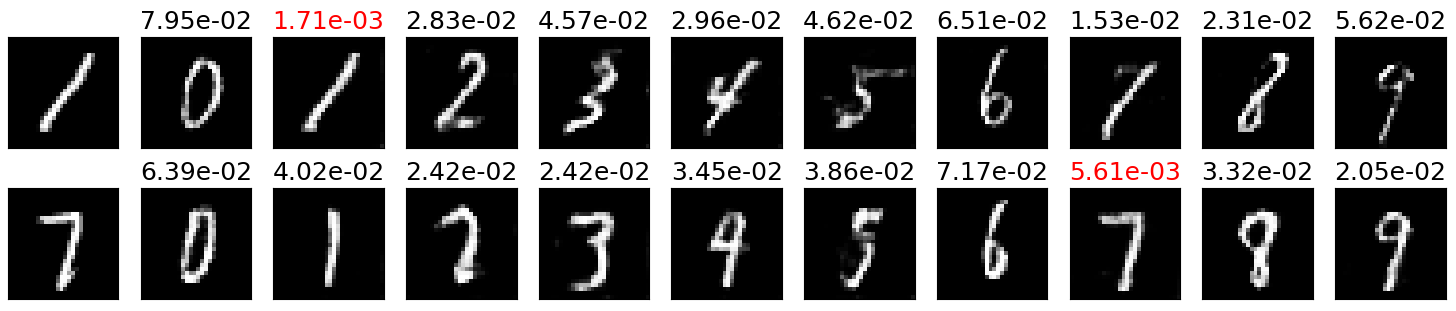}
    \caption{MNIST classification performed by a generative classifier: test image on the left accompanied with generated images and their $L^2$ distances to test point. All generators try their best to match the test image, and the predicted class is the class of the image that most closely matches the test image in $L^2$ distance.}
    \label{gc_1}
\end{figure}

Nearest neighbors has the desirable property of interpretability - with any prediction, there is a \textit{rationale} for the prediction in the form of the closest image to the test point and its metric score. By using this score as a confidence value, we can view KNN as a selective classifier, and we would expect that higher confidence thresholds would lead to more accurate predictions. Our generative classifier retains these properties - for example, we can visualize the optimization procedure, yielding a set of images that explains why our classifier made its prediction (Figure \ref{gc_1}). We can also take the maximum similarity as a confidence measure for the purpose of selective classification, and intuitively this should accurately reflect the ability of the classifier to accurately make a prediction.
\section{MNIST Experiments}

\begin{figure}[!h]
    \centering
    \begin{subfigure}{0.45\textwidth}
        \centering
        \includegraphics[scale=.6]{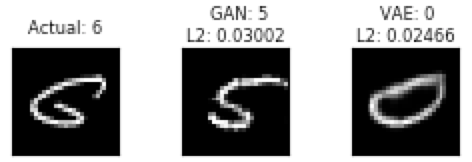}
        \caption{The GAN and VAE misclassify the 6, but their predicted classes disagree. Both generated images exhibit a right slant to match the test image.}
        \label{gc_wrong_1}
    \end{subfigure}
    \hfill
    \begin{subfigure}{0.45\textwidth}
        \centering
        \includegraphics[scale=.6]{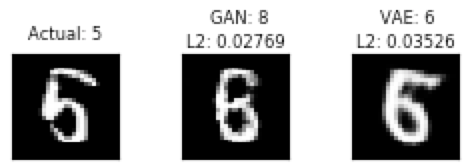}
        \caption{The 8 and 6 generators for GAN and VAE respectively are capable of producing an image similar to the 5 in $L^2$ distance.}
        \label{gc_wrong_2}
    \end{subfigure}
    \caption{The mistakes made by the generative classifier can be interpretable - on the left we see that the model has not learned how to reproduce the test image, resulting in an unconfident, incorrect prediction. On the right we see that the generator of an incorrect class is capable of reproducing the test image to some extent, pointing to degeneracies in the generator space.}
    \label{misclassification}
\end{figure}

\subsection{Model Training}
For the generative models, we trained DCGANs \citep{radford2015unsupervised} with a 15-dimensional latent space and VAEs with a 10-dimensional latent space - the dimensions were chosen by cross-validation. For both generative models, we used an $L^2$ similarity measure. To train the DCGANs, we employed techniques such as label smoothing \citep{salimans2016improved}, noise injection \citep{arjovsky2017towards}, and weight normalization \citep{salimans2016weight}, which helped to improve the quality of the images and stabilize training.
\subsection{Results}
Using a DCGAN as our generative model, we achieved an accuracy of $97.81$ percent on MNIST. Using a VAE with the encoder output in place of iterative optimization, we achieved an accuracy of 98.35 percent. As a baseline, 1 nearest neighbor achieves an accuracy of 96.91 percent.
In addition, the mistakes made by the generative classification method are readily interpretable - in Figure~ \ref{gc_wrong_1}, both GAN and VAE models are unable to produce the $6$ that is in the test set, which demonstrates that this test $6$ is dissimilar from the $6$'s seen in the training set and thus is more difficult to correctly classify. Indeed, the confidence for the GAN and VAE predictions lie within the bottom 5.48 and 7.07 percentiles of the confidences for these respective models. However, one issue with this classification method is that it highly depends on the regularity of the images that are produced by the generators - if a test image is a $5$ and the $8$ generator is capable of producing something that looks like the $5$, then it is possible for our method to misclassify the $5$ as an $8$ - see Figure~\ref{gc_wrong_2}. We observed similar results when using various recent generative model formulations that are known to produce high quality images such as Wasserstein GAN \citep{arjovsky2017wasserstein}. One possible fix is to try alternative similarity measures - the two images that are produced by the generators in Figure~\ref{gc_wrong_2} are similar to the test image in $L^2$ distance, but they may not be close with respect to another similarity measure. On the other hand, the optimization procedure can exploit more complex neural network similarity metrics by finding images that do not visually resemble the test image but still produce high similarity scores. Another approach would be to train the generators in such a way that the search space for a certain conditional generator does not contain examples from other classes. For example, BEGAN \citep{berthelot2017began} is a variation of GAN with a hyperparameter that controls the trade-off between image quality and image diversity, so it is possible that emphasizing image quality would remove such out-of-distribution examples from each conditional GAN's search space.
\begin{figure}[!h]
    \centering
    \includegraphics[scale=.5]{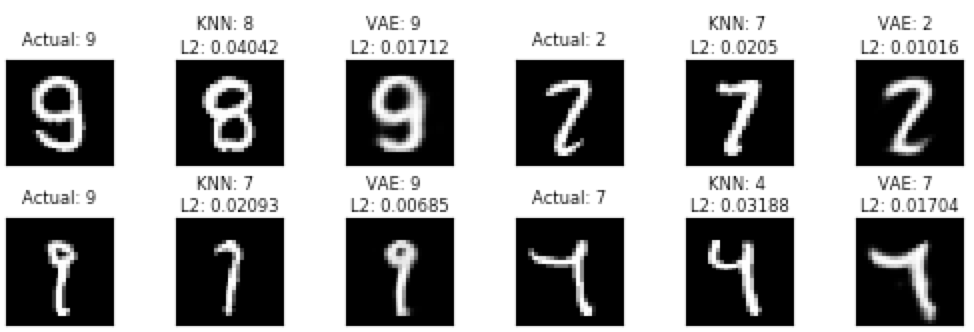}
    \caption{MNIST images misclassified by KNN but correctly classified by VAE-based GC using $L^2$ distance from test image. Each horizontal triplet of images shows the test image, the nearest neighbor in the data, and the VAE generated image. The VAE is able to produce images very similar to the test image, while no such images of the correct class exist in the training set and so KNN misclassifies the data.}
    \label{knn_wrong}
\end{figure}
\section{Out-of-Distribution Results for Generative Classifiers}
\begin{figure}[!h]
    \centering
    \begin{subfigure}{0.4\textwidth}
        \centering
        \includegraphics[scale=.4]{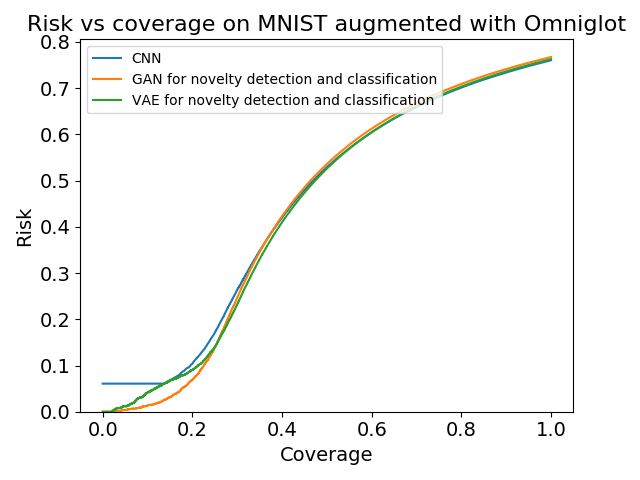}
        \caption{Using a GC as novelty detection globally outperforms using the CNN softmax threshold.}
        \label{gcclf_omniglot}
    \end{subfigure}
    \hfill
    \begin{subfigure}{0.4\textwidth}
        \centering
        \includegraphics[scale=.4]{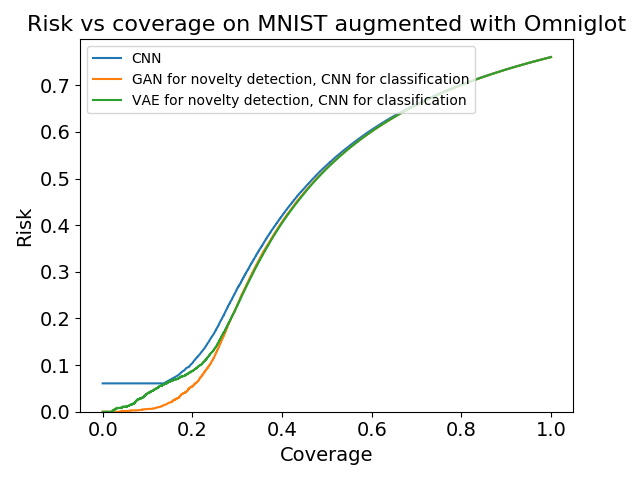}
        \caption{The GC outperforms the CNN for low values of coverage.}
        \label{gcl2_omniglot}
    \end{subfigure}
    \caption{Risk-coverage for generative classifier. Left: results for the generative classifiers (VAE and GAN) vs. the discriminative CNN. Note that for low coverage, the generative classifiers perform better (risk is lower and vanishes to zero). For higher coverage, the CNN outperforms the generative models due to its higher accuracy on in-distribution examples. Right: using the generative models for novelty detection and the CNN for classification, we obtain models that outperform the discriminative CNN for all coverage values.}
    \label{risk-coverage}
\end{figure}

\begin{figure}
    \centering
    \includegraphics[width=\textwidth]{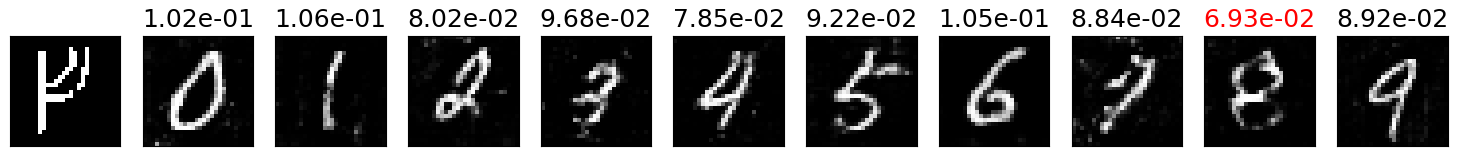}
    \caption{An Omniglot image classified with high confidence by a CNN is classified with low confidence by a GAN-based generative classifier. The $L^2$ distance to the closest image is an order of magnitude higher than that of the MNIST images in Figure~\ref{gc_1}.}
    \label{gc_2}
\end{figure}

We run the same out of distribution experiment on Omniglot-augmented MNIST using the generative classifier models described above. We find that although the generative classifier with $L^2$ distance has a lower baseline performance than the CNN on the original test set, it does possess the vanishing risk property even with the inclusion of out-of-distribution examples, as demonstrated in the risk-coverage plot in Figure {\ref{gcl2_omniglot}}. This means that in contrast to the CNN, the risk can be driven down to zero by increasing the selectivity - thus the confidence measure really does reflect the classifier's inherent ability to classify an image. As an example, Figure~\ref{gc_2} displays the optimal images produced by our generators on an Omniglot image that was classified by a CNN with the highest possible confidence. Our generators are unable to match this image, and the closest $L^2$ distance achieved is on the order of $10^{-2}$, as opposed to the $L^2$ distances on the order of $10^{-3}$ on MNIST images as seen in Figure~\ref{gc_1}. The lower confidence score of the generative classifier on Omniglot reflects the fact that we cannot actually classify these images correctly, while the high confidence of the CNN on these images misleads us to believe that we are capable of doing so.

In order to improve the accuracy on MNIST while preserving our desirable selectivity properties, we can first use the generator confidence to determine coverage -- after thresholding appropriately, we can use the CNN to make a final classification of the data. This procedure of generative novelty detection and CNN classification results in a risk-coverage curve that lies \emph{entirely} at or below the CNN curve (Figure~\ref{gcclf_omniglot}). Thus we are able to maintain the same performance on MNIST while achieving the desired selectivity properties.

Future work in improving the generative models and selecting an appropriate similarity measure may result in a generative classifier that can simultaneously outperform the CNN in classification and serve as a novelty detector.

\section{Conclusion}
We proposed a general method for classification using generative models that can be used with various VAE and GAN models. As we have shown, the generative approach offers resilience to misclassification of out-of-distribution examples, and provides a reliable measure of classification confidence. 

Much more work is required to scale our approach to more challenging image recognition domains.
At present, we are not aware of generative models that can reliably capture the latent manifold of complex realistic images such as in the Imagenet dataset~\citep{imagenet_cvpr09}, although considerable progress has been made~\citep[e.g.,~the impressive results of][]{karras2017progressive}. We believe that our work provides additional motivation for further improving the performance of generative models. Our work also offers a possible principled way of evaluating a generative model by its performance when used as a generative classifier. Future research in fundamentally understanding and controlling the behavior of generative models will increase their effectiveness when applied to problems such as safe prediction.

\section*{Acknowledgement}
This work was funded in part by Siemens and by an ONR PECASE N000141612723.

\bibliography{iclr2018_conference}
\bibliographystyle{plainnat}

\end{document}